\documentclass[conference,letterpaper]{IEEEtran}
\IEEEoverridecommandlockouts
\usepackage{comment}
\usepackage{cite}
\usepackage{amsmath,amssymb,amsfonts}
\usepackage{algorithmic}
\usepackage{graphicx}
\usepackage{textcomp}
\usepackage{xcolor}
\usepackage{array}
\usepackage{multirow}
\def\BibTeX{{\rm B\kern-.05em{\sc i\kern-.025em b}\kern-.08em
    T\kern-.1667em\lower.7ex\hbox{E}\kern-.125emX}}
\begin{document}
	
	\title{Influence of Pointing on Learning to Count: A Neuro-Robotics Model\\
	}
	
	\author{\IEEEauthorblockN{Leszek Pecyna}
	\IEEEauthorblockA{\textit{Centre for Robotics and Neural Systems} \\
	\textit{University of Plymouth}\\
	Plymouth, United Kingdom \\
	leszek.pecyna@plymouth.ac.uk}
	\and
	\IEEEauthorblockN{Angelo Cangelosi}
	\IEEEauthorblockA{\textit{Centre for Robotics and Neural Systems} \\
	\textit{University of Plymouth}\\
	Plymouth, United Kingdom \\
	a.cangelosi@plymouth.ac.uk}
	}
	
	\maketitle
	
	\begin{abstract}
	In this paper a neuro-robotics model capable of counting using gestures is introduced. The contribution of gestures to learning to count is tested with various model and training conditions.
	Two studies were presented in this article. In the first, we combine different modalities of the robot's neural network, in the second, a novel training procedure for it is proposed. The model is trained with pointing data from an iCub robot simulator.
	The behaviour of the model is in line with that of human children in terms of performance change depending on gesture production.
	\end{abstract}
	
	
	\section{Introduction}
	The ability to perceive and process quantity is undoubtedly one of the crucial skills for both humans and animals.
	
	There are many suggestions that there is a common mechanism for representing approximate numerosity in animals and humans \cite{Xu2000}. However, human quantification skills are much more advanced than those of animals. We not only estimate and perceive numerosity but we can count, i.e. enumerate objects by tagging them by numbers' names \cite{alibali1999function,wynn1992children}.
	
	The ability to count itself is not simply the recitation of numbers. Researchers found that children, although familiar with the numbers' names and their order, are not able to count objects \cite{wynn1990children,wynn1992children,le2007one}. There have been two tests introduced to investigate the development of number cognition: HM (How many) and GM (give me a number). The understanding of numbers is difficult to define; It has been assumed that, the child understands the meaning of the number when she passes a GM test \cite{wynn1992children,le2007one} and so provides number of objects that was asked for. 
	
	A significant influence of gestures in the learning process of counting  has been demonstrated in \cite{alibali1999function,fischer2012finger,graham1999role}. There is evidence that pointing and touching gestures facilitate counting accuracy in children and non-human primates \cite{alibali1999function}. They are not only used for keeping track of counted objects, but may also help with the implementation of one-to-one correspondence by contributing to the individuation of items, and thereby to the segmentation of the counting task into smaller units \cite{alibali1999function}. Such a behaviour might also provide support for working memory in the one-to-one correspondence task \cite{alibali1999function}. Furthermore, finger counting is considered to be an essential stage in number cognition \cite{Vivian2014,goldin2014gestures,soylu2018you}. Thus, there is strong evidence that embodiment has an influence on cognition, including for abstract concepts such as numbers. More about developmental robotics and embodied cognition can be found in \cite{cangelosi2015developmental}.
	
	In the field of connectionist modelling of the enumerating process, one of the first models was created by Amit in 1988 \cite{amit1988neural}. It consisted of a neural network capable of counting clock chimes. The model shows that Elman network can develop the ability to count four symbols using supervised learning with backpropagation through time. In 1999 Rodriguez et al. \cite{rodriguez1999recurrent} presented a recurrent neural network that was capable of counting letters. Another important model can be found in \cite{ahmad2001simulation} and \cite{ahmad2002connectionist}. The main focus in this model is on the distinction between sequential enumeration and subitizing (i.e. the immediate apprehension of small numerosity).
	
	Although there have been several models of the enumeration process, not many of them have been implemented on a robot platform or robot simulator. In this field Ruci\'nski et al. \cite{rucinski2011embodied}, using the iCub humanoid robot \cite{metta2010icub}, developed a model capable of number comparison and parity judgement that took into account the SNARC (spatial-numerical association of response codes) effect. In another work \cite{rucinski2012robotic} they constructed a recurrent neural network that is first taught to recite the numbers in the counting order and then to count the presented objects with or without proprioceptive gesture input from iCub robot.
	
	Finally, the influence of finger counting on number word grounding has been documented in \cite{DiNuovo2014,DiNuovo2014a,DiNuovo2015,Vivian2014,di2017embodied,shu20884}. In these studies, the iCub platform (and its simulator) was used to provide embodied conditions as well. The results support the hypothesis that learning the number words in sequence along with finger configurations helps the fast building of the initial representation of number in the robot \cite{DiNuovo2014a}.
	
	To increase embodiment, Ruci\'nski's model of learning to count \cite{rucinski2012robotic,rucinski2014modelling} has been modified and extended in the work presented here. This makes the simulations more comparable with those obtained by psychologists (e.g. \cite{alibali1999function}).
	In Ruci\'nski's work, a cognitive model of learning to count uses a recurrent artificial neural network which has as input visual stimuli and proprioceptive gesture information. The network is first pre-trained to recite numbers from 1 to 10. After achieving this skill, the network learns to stop counting when the number of elements from the visual input is reached. In this way, the neural network (NN) is counting the elements from the visual stimuli. In his work, the behaviour and results of the training of the network were compared with respect to the proprioceptive gesture information. The results showed that the learning of counting procedure is better when the gesture input is included.
	
	In the model presented in this paper, the gestures are produced by the network as an output, or there is an output-input feedback loop covering the produced gestures. Such a set-up corresponds more closely to the experiments with children in which active gestures were investigated.

	The paper is organised as follows. First, the model with its architecture and evaluation is presented. Then, the first study presenting a variety of experimental conditions is described.
	This study will help us understand the dependencies in the model and how the model internal connections influence its performance.
	The second study describes multi-stage training where the network was separately pre-trained to produce the sequence of words and a pointing signal. The article finishes with conclusions and discussion of future work.
	
	\section{Model description}
	Similarly to \cite{rucinski2012robotic,rucinski2014modelling} a recurrent neural network model based on Elman architecture \cite{elman1990finding} was used to model the process of acquisition of the counting skills. Different structures of the neural network architecture were considered in order to compare their performance, and to make the training more similar to experiments with children.
	The most complex structure can be found in Fig.~\ref{fig:architecture} where all the possible layers and connections between them are shown.
	
	The environment that was used for creating the network is the Tensorflow library and Python.	As for the robotic platform, the iCub humanoid robot and its simulator was used. The iCub is an open source humanoid robot designed to support cognitive developmental robotics research \cite{metta2010icub}. In the current state the platform is a child-like robot with 53 degrees of freedom and is 1.05 m tall. The iCub simulator has been designed to reproduce the physics and the dynamics of the physical platform. It permits the creation of realistic physical scenarios in which the robot interacts with a virtual environment \cite{DiNuovo2014a}. 
	\begin{figure}[tbp]
		\centerline{\includegraphics[width = 1\columnwidth]{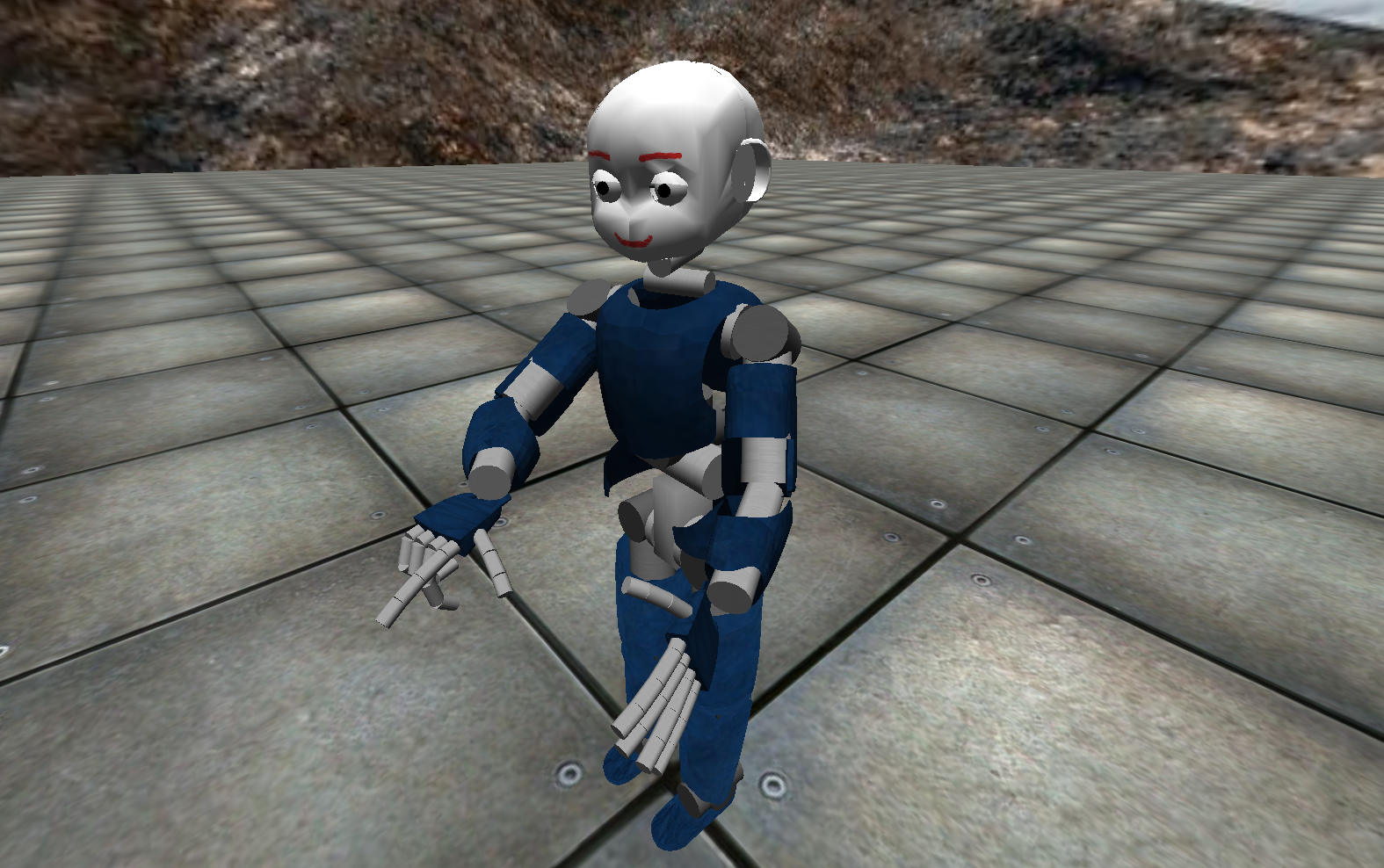}}
		\caption{The iCub simulator. Robot performing pointing.}
		\label{fig:iCub}
	\end{figure}
		\subsection{Counting task}
		
		The main task of the model is to produce a correct sequence of number words and pointing gestures. These output words (described in section \ref{model output}) correspond to the numbers from 1 to 10. The expected result is that the NN will produce the sequence of number words and stop when it reaches the last object. For example, for four objects it should produce sequence of four words corresponding to: ``one'', ``two'', ``three'', ``four''. The model was trained and tested for number of objects ranging from 0 to 10.	
		
		\subsection{Model Architecture}\label{model structure}
		\begin{figure}[tbp]
			\centerline{\includegraphics[width = 1\columnwidth]{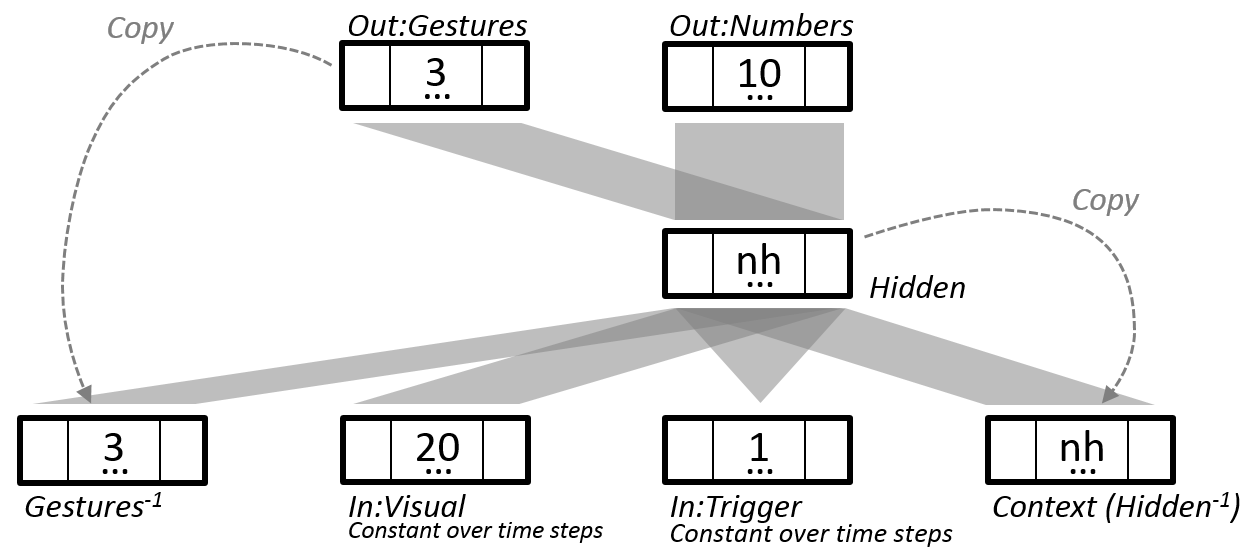}}
			\caption{Model architecture. Gray polygons represent all-to-all connections between the layers of neurons. Numbers in the rectangles corresponds to the number of neurons, ``nh'' represents the variable number of hidden units}
			\label{fig:architecture}
		\end{figure}
		The inputs and outputs of the network are organized in several blocks. Inputs: Visual and Trigger. Outputs: Gestures and Numbers (see Fig.~\ref{fig:architecture}). The blocks: Gestures\textsuperscript{-1} and Context are not considered as model inputs as they are part of the recurrent loops (they are not provided externally to the model). However, in some architectures (described in section \ref{exp_cond1}) the gestures are not produced by the model but only provided to it externally. In these cases, the gesture input will have the same structure as the one described in the section with model outputs (\ref{model output}).
		
		The network has two recurrent loops: Elman recurrent loop where the hidden units from previous time step are copied to the input units; and Jordan recurrent loop where the network copies the output block \cite{elman1990finding}.
		The architecture of  the network is very similar to the one described in \cite{rucinski2012robotic} with the difference of the additional output and second recurrent loop.

		\subsubsection{Model inputs}\label{input}
		
		\begin{itemize}
			\item Trigger:
			This is a single unit input; its role is to indicate when the counting process should start. The network is supposed to produce zeros at the output whenever the trigger unit is off. The desired counting is produced when the trigger is set to one. The trigger input value remains constant throughout the whole time sequence in the training and testing data sets.
			\item Visual: This is a 1-dimensional vector (saliency map), which can be considered a simple model of retina (same as in \cite{rucinski2012robotic}). This input consists of 20 units corresponding to 20 spacial locations. Depending on the presence of an object in a given location, each unit of the visual input will be activated or not. The maximal number of objects assumed in presented experiments is 10, their locations are chosen randomly. The visual input is normalized to 1 in order to eliminate possibility of discrimination of cardinality based on summed activation of that input.
		\end{itemize}
		
		\subsubsection{Model outputs} \label{model output}
		
		\begin{itemize}
			\item Numbers: This is a 1-dimensional vector with 10 units. For simplification and to facilitate the output and performance analysis, the one-hot coding has been used.
			This type of coding helps to calculate the percent accuracy of the network. It is, however, different from phonetic words representing the numbers, which are non-orthogonal.
			\item Gestures: The values of proprioceptive signal were collected using the iCub humanoid robot simulator. The robot was made to point to 20 locations in front of it, corresponding to 20 locations in the visual input. These locations were uniformly distributed in a line placed 15 cm above its hip and 30 cm in front of it. The length of the line was 30 cm. The right arm Cartesian controller was used to move the arm. The controlled kinematic chain consists of 6 degrees of freedom: torso yaw and pitch and the first 4 joints of the robot arm (shoulder and elbow). The torso roll angle was disabled to avoid unnatural postures. The joint angles were stored. Additionally, the arm base position (where the iCub arm is down aside its body) was also included. The Principle Component Analysis has been applied on the  data and the first 3 principal components were selected. This reduced the dimensionality of the original signal from 6 to 3. These 3 components carry more than 97\% of the total statistical information of the original data.
			
		\end{itemize}
		
		\subsection{Evaluation of the results}\label{evaluation}
		Similar to \cite{rucinski2012robotic}, the intention in the model design was to perform experiments inspired by those with child participants. More precisely, the study presented in \cite{alibali1999function} - where the function of gestures in learning to count was investigated - was used to build this model and counting task. In this text we present the training for a variety of network configurations and we discuss whether it can be compared to human learning.
		
		To obtain the results presented in this article for each type of configuration, at first a hyperparameter search was performed to chose the learning rate value and hidden layer size.
		It has been observed that very often the size of a hidden layer in some range did not influence the result much. Thus, the final choice of the learning rate and hidden layer size was made in such a way that those parameters where the same over different experimental conditions. The reason for this was to make the results comparable. Their values will be presented separately for each study (see sections \ref{combination-training} and \ref{double-training})
		
		After these parameters where chosen, the model was trained in 15 independent repetitions in order to verify the robustness of the results and obtain reliable standard deviation and mean value of them. For each of these 15 training runs, the network has been tested with 50 different test sets to obtain the mean value of the results for each of the training processes.
		The values of the training data and test data were generated as separate sets. A random generator has been used to obtain them: the training order of numbers was randomised (from 1 to 10) and the spacial locations of the objects, out of 20 possible positions (note that possible arrangement of the objects grows exponentially with their number up to 6.7E11 for 10 objects).
		
	\section{Study 1: combination of modalities} \label{experiment1}
	
	The intention of this study is to investigate a variety of the training configurations that can be obtain by the model. This will help to comprehend the dependencies in the model and what kind of inputs and outputs have significant influence to the model performance.
	
		\subsection{Experimental Conditions}\label{exp_cond1}
		
		All possible combinations of inputs and outputs together with their correlations with human trials are presented below. Several configurations were omitted as they had no psychological reference e.g. training with only gestures as an input and as output. The results for listed conditions will be presented in Table~\ref{tab1}.
		
		Some of these experimental conditions are similar to those presented in \cite{rucinski2012robotic}. There are however, some differences in the training procedure as described in section \ref{combination-training}. All of the presented configurations used trigger input.
		
			\subsubsection{Input: Visual; Output: Numbers}
			Similar to the training shown in \cite{rucinski2012robotic}. This configuration corresponds to the condition when a child is only counting the objects since is not allowed to point to them. This experimental condition could be obtained by the network as shown in Fig.~\ref{fig:architecture}, where Gestures output and input blocks (together with output-input feedback loop between them) were removed.
			\subsubsection{Input: Visual; Output: Gestures}
			The purpose of this condition was to check the hypothesis that gesture proprioceptive signals might be easier to learn in a separate dedicated training, when no counting is performed. This configuration corresponds to the situation when a child is asked to point to all objects from left to right. The network structure is very similar to the previous experimental condition with the only difference that the numbers' output is replaced by the gestures' one.
			Although training seems to be similar, there is a direct correlation between visual input and gestures (gesture signal exactly corresponds to randomly chosen visual position). There is no such direct correlation between number's tag and visual input e.g. the object in position 7 can be a second object from the left (have a number's tag 'two'). Thus, we presumed such training might be easier for the network and can give interesting results.
		
			\subsubsection{Input: Visual; Output: Numbers and Gestures}
			This training configuration is similar to the next one. However, the output-input feedback loop was not used (together with gesture input block, which was a part of that loop).
			Such training corresponds to the situation where the child is producing the gestures herself and at the same time she counts the presented objects.
			
			\subsubsection{Input: Visual (and Gestures from a loop); Output: Numbers and pointing}
			This configuration is depicted in Fig.~\ref{fig:architecture}. In this case the network uses all its inputs and outputs.
			We consider this training (as well as the previous one) to correspond to the situation where the child is producing the gestures herself while counting the presented objects.
			Pointing to the object is produced together with the number word. At the next step (next object), such proprioceptive gesture signal from the previous pointing is used as an input.
			Although, this condition seems more complete, it is also more complicated and might be more difficult to train. Moreover, the discrete character of time (causing the one time step difference between gesture input and output) might implicate that it is not as adequate to the child pointing case as we intended it to be.

			\subsubsection{Input: Gestures; Output: Numbers}
			This training corresponds to a situation when the child's hand is guided by the teacher and they are pointing to the object, but the child eyes are covered so she cannot see the objects.
			\subsubsection{Input: Gestures; Output: Numbers and Gestures}
			This is very similar training to the previous condition, the only difference is that the gesture signal is additionally produced by the network (there is no output-input feedback loop in this case). This training is intend to show if such a passing of pointing signal through the network could give any benefits for the training (compare to gestures used as only input or only output cases).
			\subsubsection{Input: Visual and Gestures; Output: Numbers} \label{input_gest}
			A similar experimental configuration as used in \cite{rucinski2012robotic}. In this condition, the proprioceptive signal from gestures was given as an input to the network together with the visual input (and trigger). Ruci\'nski compared this to a situation where the pointing is performed by a puppet, as described in \cite{alibali1999function}, and presented to a child that is required to count (without performing pointing herself). As the nature of the gesture signal is proprioceptive and not visual, we would rather consider this case more similar to the situation where the child hand is held and the pointing is performed by an external teacher.
			\subsubsection{Input: Visual and Gestures; Output: Numbers and Gestures}
			This configuration is similar to the condition described previously with a gesture signal additionally produced by the network.

		\subsection{Model Training} \label{combination-training}
		The network is trained to count the objects presented in the form of visual input and/or in the form of proprioceptive gesture signal. The training uses supervised learning by backpropagation through time. Differently to \cite{rucinski2012robotic}, the results were obtained using one stage training - no pre-training was used.
		The training data set consists of two sequences of length 12.
		The visual input for both these sequences was the same; it was constant over the 12 time steps and it was representing the positions of objects in front of the robot. In the first sequence, the trigger input, all number target outputs and target and/or input gestures were equal to zero throughout all time steps. For the second sequence the trigger input was equal to one, the number targets consisted of a sequence of words corresponding  to the numbers (from 1 to up the number of objects presented to the model) and the remaining words of the sequence were set to zero. Similarly, target and/or input gesture sequence consisted of proprioceptive signal corresponding to the locations of presented objects, from left to right. The remaining positions in the sequence were consequent repetition of the last pointing value. This approach for gestures is reasonable when they are given as an input; when a zero value or base position is set in the remaining elements of the sequence, the NN easily learns to distinguish it and stop counting when that value appears (reaching 100\% accuracy). 
		
		Such a collection of two sequences was constructed for each of 11 possible numbers (possible amount of objects) from 0 to 10 (giving 22 sequences, each with 12 times steps). We will call this set a sub-epoch set, the order of numbers (represented by pair of sequences) was chosen randomly with an alternating trigger. Training sub-epoch and test sub-epoch sets were randomly generated; each of them consists of all the same numbers from 0 to 10 but each of these numbers have objects in random, different spacial locations (for visual input and gestures). 
		
		The Adam optimizer was used for the training. Each training cycle consists of 20000 sub-epochs, using batch training (the batch consists a full 22 element sub-epoch set). A starting learning rate was chosen to be 0.005 and a hidden layer size was set to 68 (chosen in a way described in section \ref{evaluation}). An exception was made for a condition 2 (Inputs: V, Outputs: G) and the learning rate was set to 0.02 in order to obtain good results. The weights were initialized using Glorot uniform initializer and biases were initialized with zero values.
		
		The counting numbers are considered as correct when the whole sequence for a given number of objects was fully correct.
		For the gesture output, the sequence could be partially correct when some of the sequence elements were correct (e.g when counting 10 elements and the pointing was performed correctly to 3 of them, the performance for that sequence would be 30\%) . As the remaining elements of the sequence are constant and are less important from the point of view of the correct counting (as they are repetition of the last element) we will only consider n+1 first elements of the sequence, where n is the number of counted objects.

		\subsection{Results}
		\begin{table}[htbp]
			\caption{Collected results for Study 1}
			\begin{center}
				\begin{tabular}{>{\centering}p{0.1\columnwidth}
						>{\centering}p{0.14\columnwidth}
						>{\centering}p{0.14\columnwidth}
						>{\centering}p{0.2\columnwidth}
						>{\centering\arraybackslash}p{0.2\columnwidth}}
					\hline
					\multicolumn{3}{c}{\textbf{Experimental Conditions}}&\multicolumn{2}{c}{\textbf{\% of correct responses$^{\mathrm{a}}$}}\\
					\textbf{Cond.}&\textbf{Inputs}&\textbf{Outputs}&\textbf{Counting$^{\mathrm{b}}$}&\textbf{Gestures} \\
					\hline
					1)&V&N&90.3 (4.9)&-\\
					\hline
					2)&V&G&-&60.4 (8.7)\\
					\hline
					3)&V&N, G&86.8 (6.8)&34.2 (2.9)\\
					\hline
					4)&V (+ G$^{\mathrm{c}}$)&N, G&84.9 (8.1)&32.5 (4.0)\\
					\hline
					5)&G&N&97.5 (1.8)&-\\
					\hline
					6)&G&N, G&96.9 (1.6)&97.7 (1.9)\\
					\hline
					7)&V, G&N&96.8 (3.8)&-\\
					\hline
					8)&V, G&N, G&98.6 (1.2)&92.0 (4.4)\\
					\hline
					\multicolumn{5}{l}{$^{\mathrm{a}}$Values are presented in a form: mean (SD).}\\
					\multicolumn{5}{l}{$^{\mathrm{b}}$\% of correct counting responses for 1 to 10 items.}\\
					\multicolumn{5}{l}{$^{\mathrm{c}}$Gestures from a loop.}
				\end{tabular}
				\label{tab1}
			\end{center}
		\end{table}

		The results of all simulations are presented in Table~\ref{tab1}. The results for counting are higher than those presented in \cite{rucinski2012robotic}. This is especially visible in case of experimental conditions 1 (with visual input and numbers as an output). There are several reasons for this. The main one depends on the usage of the Adam optimizer, which was able to increase the performance of the network and learn counting much better even when only visual input was given. Additionally, the improvement of the training was achieved by using batch training (with sub-epochs sets used as batches). When gradient descent optimizer was used we obtained results much more similar to \cite{rucinski2012robotic}
		
		In our study the results were not much improved when pre-training of number recitation was used (as it was in \cite{rucinski2012robotic}). The reason for this might be that they were already very high. For example for the configuration 7 (inputs: V,G and outputs: N) we obtained 96.8\% (Table~\ref{tab1}) and when pre-training of numbers (Stage 1A from Study 2) was used, additionally, we got a result of 97.6\%. Thus, those results were not presented.
		
		All of the repetitions were separate simulations giving a final neural network tested with test data. A one-way ANOVA was used between  groups of results for statistical analysis.
		
			\subsubsection{Counting results}
			As expected, and described in \cite{rucinski2012robotic}, the results with gesture input were significantly higher than those without them.
			When we compare the corresponding results where the gestures were added as an input: the analysis showed a strong significant difference when condition 1 with 7 and 3 with 8 were compared (${p<0.001}$ in both cases). It was also visible that using gesture input instead of visual, gave strong significant improvement as well (comparing 1 with 5 and 3 with 6 gave ${p<0.001}$).
			Surprisingly, there was no improvement of counting observed when the network was trained to produce gestures and used them in an output-input feedback loop. The results for such training were worse than those where only visual input was provided (see condition 1 and 4 in Table~\ref{tab1}). What was also unexpected, the configuration with the output-input feedback loop (4\textsuperscript{th} condition) performed worse than the one without the additional gesture loop (3\textsuperscript{rd} condition).
			However, the difference was not statistically significant ($p = 0.5$).
			As mentioned before, the reason why results with a loop are not better might be that the model is more complicated and needs more training time or it might be because of time step delay appearing in the output-input loop. When gestures were used as an independent input and output their influence on counting (compare to the case where gestures were only the input) was not visible. 
			
			\subsubsection{Gestures results}
			Important observation from the results is that the pointing performance was much better when the network was trained only to produce gestures without being trained to count - the percentage of correct responses was almost doubled (see conditions 2,3 and 4 in the Table~\ref{tab1}). Obviously, the configuration with gesture inputs gave much better gesture results, as the network have the full information at the input (condition 6 and 8).
		
	\section{Study 2: double pre-training}\label{double}
	As the results from the Study 1 did not provide some of the expected improvement of the network performance, another experiment was proposed where a more complex multi-stage training was introduced. In this study we check the performance of learning to count and to produce gestures when one or two pre-training stages were added (for number recitation like in \cite{rucinski2012robotic} and for pointing). 

		\subsection{Experimental Conditions}\label{exp_cond2}
		There are only two types of experimental conditions considered in this study (both discribed in section \ref{exp_cond1}). They both can correspond to a situation where a child is producing gestures herself and counts presented objects at the same time:
		\begin{itemize}
			\item Input: Visual (and Gestures from a loop); Output: Numbers and pointing
			\item Input: Visual; Output: Numbers and Gestures
			
		\end{itemize}
		
		\subsection{Model Training}\label{double-training}
\begin{figure}[tbp]
			\centerline{\includegraphics[width = 1\columnwidth]{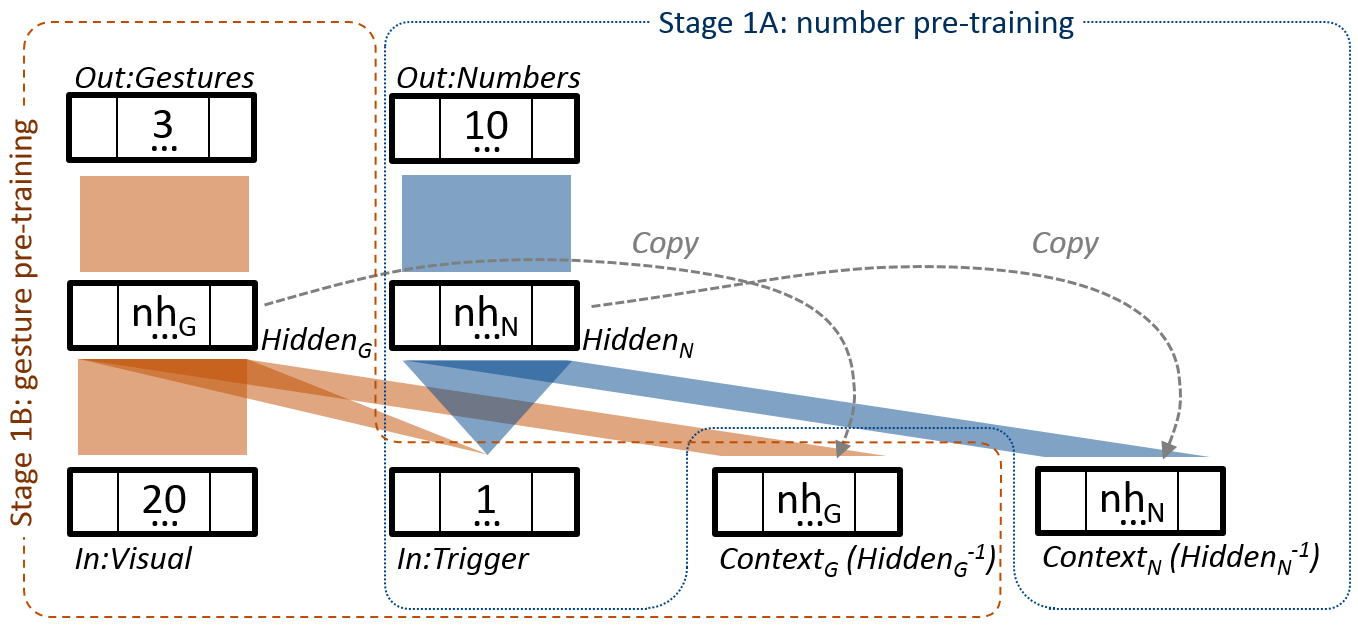}}
			\caption{Double pre-training. Polygons represent all-to-all connections between the layers of neurons. Numbers in the rectangles corresponds to the number of neurons, ``nh'' represents their number in the hidden layer}
			\label{fig:double}
		\end{figure}

		The training of this model can be divided into three stages, two pre-trainings (Stage 1A and Stage 1B) and the main training (Stage 2). The architecture of the trained networks from both pre-trainings can be found in Fig.~\ref{fig:double}. These two stages could be run in parallel as they use different weights (different parts of the NN).
		We considered and compared simulations without pre-training, with one or with both pre-trainings.
		
		All the training stages used a supervised learning by backpropagation through time with the Adam optimizer.
		
		The training described below was performed in two training conditions where a sequence of pointing proprioceptive signals was constructed in a different way. The first condition was the same as described in Study 1:
		\begin{itemize}
			\item ``Stay at the last one'': For the trigger input set to zero, the input gestures were equal to zero throughout all time steps. For the trigger input set to one, input gestures consisted of a sequence corresponding to the locations of the presented objects, from left to right, and the remaining gestures in the sequence were consequent repetition of the last pointing value. As mentioned before, this approach is reasonable when gestures are given as an input. This condition was used in the second study to make the results comparable with Study 1.
			\item ``Go to base'' (introduced one): When the trigger was equal to zero, the input gestures were set to the base position signal throughout all time steps. The base position was obtained when the robot arm was directed down the robot body (the iCub ``home position'' for the robot simulator).
			When the trigger was equal to one, the input gestures consisted of sequence corresponding to the locations of presented objects, from left to right (as before). The remaining gestures were set to the base position value. This condition seems to be much more similar to a human condition, where a child simply stops pointing after the counting is finished.
		\end{itemize}
			\subsubsection{Stage 1A number pre-training}
			Psychological studies shows that human children can recite numbers, so they acquire a list of tags (numbers' names), prior to acquisition of counting skills \cite{gelman1980young}. In this training stage the network is taught to recite the numbers' names from 1 to 10 (same type of pre-training as in \cite{rucinski2012robotic}). In this stage, neural network uses only the trigger as an input (see Fig.~\ref{fig:double}). Similarly as in Study 1, the training data set consist of two sequences of length 12. In the first sequence, the trigger input and all number target outputs were set to zero throughout all the time steps. In the second sequence, the trigger input was one and the number targets consist of a sequence of words corresponding  to the recited numbers, from 1 to 10 and the remaining 2 words were set to zero.
			
			The training was run for 7000 epochs with a learning rate of 0.01 and a hidden layer with 20 neurons and the weights were updated using batch training. The size of the hidden layer and learning rate were chosen using the same approach as described in section \ref{evaluation}.
			\subsubsection{Stage 1B gesture pre-training}
			When children start to count (around 2-3 years old \cite{le2007one}) they are already capable of complex motor skills (children successfully perform reaching task before their age of 28 months \cite{bertenthal1998eye}). To reflect this in our studies, we decided to add another pre-training where the network is taught to point correctly to the locations indicated by the objects in the visual input.  In this stage, the neural network uses visual and trigger inputs and produces a gesture sequence as an output (see Fig.~\ref{fig:double}).
			This training stage is conducted in the same manner as the training in the experimental condition number 2 in the first studies (with additional ``go to base'' condition being applied).

			\subsubsection{Stage 2 learning to count}
			This training is run in exactly the same manner as Study 1's trainings. With the remark that two training conditions were considered.  However, the learning rate was again chosen in a manner described in section \ref{evaluation} and set to 0.001. The hidden layer was a sum of hidden layer neurons from Stage 1A and 1B. We found that results for Stage 1A were practically perfect for a wide range of hidden layer size, so the size used in that stage (20 neurons) was chosen to optimize the performance of the whole multi-stage training. As mentioned before this training was conducted for two configurations with the output-input loop and without it (see section \ref{exp_cond2}). The network is fully connected, thus, it can be found that some of the weights were not pre-trained. Their values were initialized using Glorot uniform initializer.
		
		\subsection{Results}
		\begin{table}[tbp]
			\caption{Collected results for Study 2 - Counting}
			\begin{center}
				\begin{tabular}{>{\centering}p{0.05\columnwidth}
						>{\centering}p{0.05\columnwidth}
						>{\centering}p{0.15\columnwidth}
						>{\centering}p{0.15\columnwidth}
						>{\centering}p{0.15\columnwidth}
						>{\centering\arraybackslash}p{0.15\columnwidth}}
					\hline
					&&\multicolumn{4}{c}{\textbf{Pre-trainings used$^{\mathrm{a}}$}}\\
					&&\textbf{No}&\textbf{Stage 1A}&\textbf{Stage 1B}&\textbf{Both}\\
					\hline
					\multirow{2}{*}{\textbf{S$^{\mathrm{b}}$}}&\textbf{L$^{\mathrm{d}}$}&84.9 (8.1)&87.9 (5.9)&95.0 (2.4)&97.5 (1.3)\\
					\cline{2-6}
					&\textbf{NL$^{\mathrm{e}}$}&86.8 (6.8)&88.5 (5.2)&95.9 (2.6)&97.8 (1.0)\\
					\hline
					\multirow{2}{*}{\textbf{B$^{\mathrm{c}}$}}&\textbf{L}&85.0 (11.7)&89.7 (5.4)&99.1 (1.4)&99.6 (0.4)\\
					\cline{2-6}
					&\textbf{NL}&87.4 (7.9)&87.4 (6.4)&99.0 (0.9)&99.5 (0.6)\\
					\hline
					\multicolumn{6}{l}{$^{\mathrm{a}}$\% of correct responses for 1 to 10 items, in a form: mean (SD).}\\
					\multicolumn{6}{l}{$^{\mathrm{b}}$``Stay at the last one'' training approach.}\\
					\multicolumn{6}{l}{$^{\mathrm{c}}$``Go to the base'' training approach.}\\
					\multicolumn{6}{l}{$^{\mathrm{d}}$Training with output-input gesture loop.}\\
					\multicolumn{6}{l}{$^{\mathrm{e}}$Training with no output-input gesture loop.}
					
				\end{tabular}
				\label{tab2}
			\end{center}
		\end{table}
	\begin{table}[tbp]
		\caption{Collected results for Study 2 - Gestures}
		\begin{center}
			\begin{tabular}{>{\centering}p{0.05\columnwidth}
					>{\centering}p{0.05\columnwidth}
					>{\centering}p{0.15\columnwidth}
					>{\centering}p{0.15\columnwidth}
					>{\centering}p{0.15\columnwidth}
					>{\centering\arraybackslash}p{0.15\columnwidth}}
				\hline
				&&\multicolumn{4}{c}{\textbf{Pre-trainings used}}\\
				&&\textbf{No}&\textbf{Stage 1A}&\textbf{Stage 1B}&\textbf{Both}\\
				\hline
				\multirow{2}{*}{\textbf{S}}&\textbf{L}&32.5 (4.0)&33.0 (2.9)&43.0 (2.9)&64.8 (2.3)\\
				\cline{2-6}
				&\textbf{NL}&34.2 (2.9)&34.8 (2.1)&45.3 (2.4)&65.1 (3.1)\\
				\hline
				\multirow{2}{*}{\textbf{B}}&\textbf{L}&24.2 (2.2)&25.2 (2.5)&44.2 (2.4)&56.5 (2.3)\\
				\cline{2-6}
				&\textbf{NL}&26.7 (2.8)&25.4 (1.8)&45.2 (2.8)&56.9 (2.2)\\
				\hline
			\end{tabular}
			\label{tab3}
		\end{center}
	\end{table}
	\begin{figure}[tbp]
		\centerline{\includegraphics[width = 1\columnwidth]{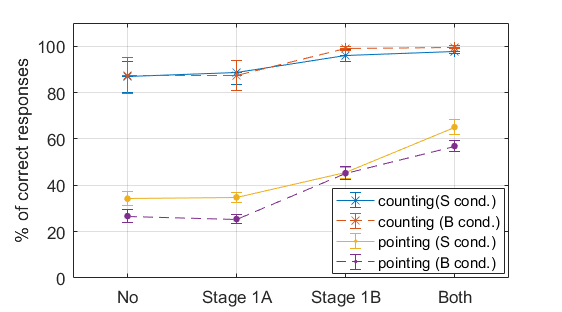}}
		\caption{Results for different pre-training options, for the network with no gesture loop and for both training conditions (S and B).}
		\label{fig:plot_pre}
	\end{figure}
	The model was trained in 15 independent repetitions, where each of the training cycles (repetitions) finished with testing the network with 50 different test sets. The mean values of the results and their standard deviations (in brackets) are shown in Table~\ref{tab2} and \ref{tab3}. For statistical analysis a one-way ANOVA was used to compare each of the presented results with one another.
	\subsubsection{Counting results}
	As can be seen in the tables and in Fig.~\ref{fig:plot_pre} the best results were obtained when both pre-trainings were applied and the worst when no pre-training was used, as expected.
	Statistical analysis in all 4 conditions (S+L, S+NL, B+L, B+NL) between different pre-training options was performed:
	\begin{itemize}
		\item with no pre-training and with Stage 1A, showed that there is no significant difference.
		\item with no pre-training and with Stage 1B, showed that there is a strong significant difference (${p < 0.001}$ for all conditions)
		\item with Stage 1A pre-training and with both pre-trainings, showed that there is a strong significant difference (${p < 0.001}$ for all conditions)
		\item with Stage 1B pre-training and with both pre-trainings, that there is a significant difference for the training condition ``stay at the last one'' but no significance for ``go to base'' ($p = 0.0017$, $0.02$, $0.19$, and $0.11$ respectively)
	\end{itemize}

	As described above, Stage 1B had a bigger positive impact on the training with respect to Stage 1A.
	\begin{figure}[tbp]
		\centerline{\includegraphics[width = 1\columnwidth]{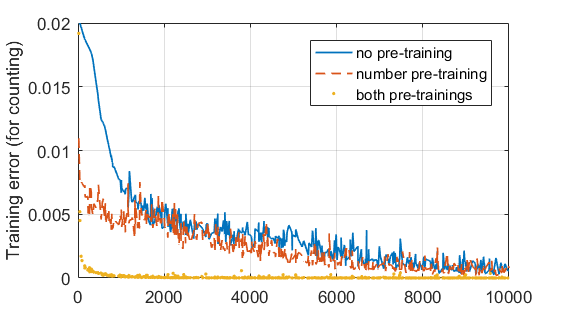}}
		\caption{Typical example of Stage 2 training loss functions (counting part of it) for first 10000 sub-epochs, for trainings with different pre-training options.
		Plots obtained for training with no gesture loop and for ``Go to base'' condition.}
		\label{fig:no_both}
	\end{figure}
	However, as shown in Fig~\ref{fig:no_both}, the influence of number pre-training is easily noticeable in the first stages of the training.
	
	As in Study 1, there is no impact of using the gesture loop in the training. For all four pre-training options in both ``stay at the last one'' and ''go to base'' conditions the difference between training with the loop or without it was statistically not significant ($p \in [0.32, 0.85]$ for all 8 considered pairs).
	
	It was expected that the network will give better results when trained in the ``go to base'' condition. As mentioned before, in this case the neural network could potentially recognise the situation when the base position is achieved to stop counting. This influence is visible when Stage 1B pre-training or both pre-trainings were used. Analysis showed that for these training options (and for network with or without gesture loop) the difference (between S and B conditions) is strongly significant (${p < 0.001}$). This influence was not found when gestures' pre-training was not used (``No'' and ``Stage 1A'' pre-training options).

	\subsubsection{Gestures results}
	Similarly, as in case of counting the best performance of the network was observed when both pre-trainings were applied and the worst when no pre-training was performed.
	Statistical analysis for all 4 conditions (S+L, S+NL, B+L, B+NL) between pre-training options was performed for gestures results:
	\begin{itemize}
		\item with no pre-training and with Stage 1A, showed that there is no significant difference.
		\item with no pre-training and with Stage 1B, showed that there is a strong significant difference (${p < 0.001}$ for all conditions)
		\item with Stage 1A pre-training and with both pre-trainings, showed that there is a strong significant difference (${p < 0.001}$ for all conditions)
		\item with Stage 1B pre-training and with both pre-trainings, showed that there is a strong significant difference (${p < 0.001}$ for all conditions)
	\end{itemize}
	The influence of number recitation pre-training was not visible when comparing the training using Stage 1A pre-training with the training where no pre-training was performed. However, the number recitation pre-training had a very significant positive impact on the results when both pre-trainings were used (in comparison to the situation where only gesture pre-training was applied). This impact is
	stronger than expected for gesture results (stronger than observed in case of counting results).
	
	The difference of performance between networks with gesture loop and without was again not observed; it was statistically significant only in two configurations (S with Stage 1B pre-training: $p = 0.036$ and for condition B when no pre-training was performed: $p = 0.016$) in favour of the network without the loop.

	``Stay at the last one'' training gave better pointing results, with the exception of trainings where only Stage 1B pre-training was used. A one-way ANOVA analysis showed a strong significance of this difference for those conditions (${p < 0.001}$).
	This gesture trend is already visible in Stage 1B pre-training results. Those partial results are as follows:
	\begin{itemize}
		\item ``Stay at the last one'', 60.4\% (8.7)
		\item ``Go to base'', 45.3\% (4.0)
	\end{itemize}
	 It seems that in case of the training with gesture pre-training, the neural network loses some of its capability to point in favour of counting\footnote{The training loss function is calculated as a sum squared error. This definition of loss function and the fact of using gestures as real numbers cause that their weight in the final result is, in some way, lower than the counting weight}. This is not visible for other pre-training options, possibly because, in other cases there is a dedicated (by applying number pre-training) part of the network to produce numbers and so the part responsible for gestures is in some way isolated. This might be also the reason why results for training with only Stage 1B pre-training are significantly worse than those where both pre-trainings were used.
	
	\section{Conclusions and discussion}
	
	In this paper a neuro-robotics model with recurrent artificial neural network capable of producing the pointing gestures while counting the objects was proposed. The model allows us to measure the contribution of the produced pointing signal to learning to count.
	The presented studies show a high improvement of counting when the network is pre-trained to perform pointing and then use them while counting.
	Results confirms that the generation of gestures improves the model's counting abilities.
	
	Moreover, the results are not only better than the one obtained by the network with only visual input but comparable or even better than the results where the network uses gestures as an input.
	When we compare the results from the network with gesture input (configuration 7 in Study 1) pre-trained for recitation in ``stay at the last one'' training condition with results from NN from Study 2 (with both pre-trainings) in ``go to base'' condition we get the results 97.6\% and 99.5\% respectively. The improvement is in favour of the network with gestures produced as an output (ANOVA analysis showed statistical significance, $p = 0.0033$).
	When ``stay at the last one'' condition was used for both networks we obtained very similar results (97.6\% and 97.8\%).
	This observation stays in line with the human study presented in \cite{alibali1999function}.
	
	When multi-stage training was performed, the learning rate values were chosen using hyper parameter search for each stage of the training to optimize the results. Their values in pre-trainings were higher then the one found for the final training. This is also in accordance with neuroscience studies \cite{johnston2001sculpting}.
	
	A few issues concerning the model are discussed below. Due to the RNN implementation of discrete time the synchrony between gestures and number words produced is easily obtained. However, in the case of children, synchronising the number words with counted objects may be one of the major functions of gestures \cite{alibali1999function,graham1999role}. To address this issue, a model with continuous time would have to be considered.
	
	Another possible improvement of the model could be achieved by using number words more similar to names used in human language. Such a change could potentially influence the sequence of words produced by the network (e.g. if two numbers have similar names they could be swapped causing a string error).
	
	Finally, a more complex visual input could be used. This would possibly require significant extension of the network. In this situation, a deep neural network with spatial filters and partially unsupervised training  might be implemented, as applied for numerosity visual sensation and numerosity comparison in \cite{stoianov2012emergence}.
	
	The results presented in this article provide quantitative evidence in support of cognitive embodiment theory that number cognition and counting can be boosted by gestures. It seems clear that the network correlated the spatial position of the objects (and pointing procedure) with the numbers' tags, as otherwise it would not be able to take advantage of the produced gestures. The future, analysis of the internal behaviour of the model can be valuable in understanding of the contribution of pointing on learning to count.
	
	\bibliographystyle{IEEEtran} 
	\bibliography{ref}

\end{document}